\documentclass{article}
\usepackage{times}
\usepackage{epsfig}
\usepackage{graphicx}
\usepackage{amsmath}
\usepackage{amssymb}
\usepackage{amsfonts} 
\usepackage[normalem]{ulem}
\usepackage[table,xcdraw,dvipsnames]{xcolor}
\usepackage{colortbl}
\usepackage{color,soul}  % For highlighting
\definecolor{ao(english)}{rgb}{0.0, 0.5, 0.0}
\def\etal{\emph{et al.\ }}

\usepackage[final,nonatbib]{neurips_2022}

\usepackage[utf8]{inputenc} % allow utf-8 input
\usepackage[T1]{fontenc}    % use 8-bit T1 fonts
\usepackage{hyperref}       % hyperlinks
\usepackage{url}            % simple URL typesetting
\usepackage{booktabs}       % professional-quality tables
\usepackage{amsfonts}       % blackboard math symbols
\usepackage{nicefrac}       % compact symbols for 1/2, etc.
\usepackage{microtype}      % microtypography
\usepackage{xcolor}         % colors

\title{Video based Object $\mathbf{6}$D Pose Estimation using Transformers}

\author{%
  Apoorva Beedu \\
  Georgia Institute of Technology\\
  \texttt{abeedu3@gatech.edu} \\
   \And
   Huda Alamri \\
  Georgia Institute of Technology \\
  \texttt{halamri@gatech.edu} \\
   \AND
   Irfan Essa    \\
  Georgia Institute of Technology \\
  \texttt{irfan@gatech.edu} \\
}

\begin{document}

\maketitle

%%%%%%%%% ABSTRACT
\begin{abstract}
We introduce a Transformer based $6$D Object Pose Estimation framework \textit{VideoPose}, comprising an end-to-end attention based modelling architecture, that attends to previous frames in order to estimate accurate $6$D Object Poses in videos. 
Our approach leverages the temporal information from a video sequence for pose refinement, along with being computationally efficient and robust. %,  to support robotic and AR domains. 
Compared to existing methods, our architecture is able to capture and reason from long-range dependencies efficiently, thus iteratively refining over video sequences.
Experimental evaluation on the YCB-Video dataset shows that our approach is on par with the state-of-the-art Transformer methods, and performs significantly better relative to CNN based approaches. 
Further, with a speed of 33 fps, it is also more efficient and therefore applicable to a variety of applications that require real-time object pose estimation. Training code and pretrained models are available at \href{https://github.com/ApoorvaBeedu/VideoPose}{https://github.com/ApoorvaBeedu/VideoPose}.

%\irfan{Add here, why is this useful, or the benefit!} 

\end{abstract}

%-------------------------------------------------------------------------
\section{Introduction}
\label{sec:intro}
Estimating the $3$D translation and $3$D rotation for every object in an image is a core building block for many applications in robotics~\cite{saxena2008robotic,Tremblay2018DeepOP,eppner2019billion} and augmented reality~\cite{marchand2015pose}.
%that involves interactions with objects. 
The classical solution for such $6$-DOF pose estimation problems utilises a feature point matching mechanism, followed by Perspective-n-Point (PnP) to correct the estimated pose~\cite{Rad2017BB8AS,tekin2018real,peng2019pvnet,hu2019segmentation}.
However, such approaches fail when objects are texture-less or heavily occluded. 
Typical ways of refining the $6$DOF estimation involves using additional depth data~\cite{wang2019densefusion,Brachmann2014Learning,hinterstoisser2012model,konishi2018real} or post-processing methods like Iterative Closest Point (ICP) or other deep learning based rendering methods ~\cite{xiang2018posecnn,Kehl2017SSD6DMR,li2018deepim,song2020hybridpose}, which increase computational costs. Other approaches treat it as a classification problem~\cite{tulsiani2015viewpoints,Kehl2017SSD6DMR}, resulting in reduced performance as the output space is not continuous.

In robotics, augmented reality, and mobile applications, the input signals are typically videos rather than single images, thus, giving opportunity for a multi-view framework. 
Li~\etal\cite{li2018unified} utilize multiple frames from different viewing angles to estimate single object poses.
Wen~\etal\cite{wen2020se} and Deng~\etal\cite{ deng2019poserbpf} use tracking methods to estimate the poses, however these methods do not explicitly exploit the temporal information in the videos. The idea of using more than one frame to estimate object poses has seen limited exploration. As the object poses in a video sequence are implicitly related to camera transformations and do not change abruptly between frames, and as different viewpoints of the objects aid in the pose estimation ~\cite{labbe2020cosypose, chen2017multi}, we believe that modelling temporal relationship can only aid in effective perfomance on the task. 
% Extending this approach, \cite{li2018deepim,xiang2018posecnn} introduce post-processing mechanisms to improve the estimation which results in additional computational cost.  \irfan{all of this seems like related work then real motivation!}

Motivated by this, we introduce a video based object $6D$ pose estimation framework, that uses past estimations to bound the $6D$ pose in the current frame. Specifically, we leverage the popular Transformer architecture~\cite{vaswani2017attention,GPT2} with causal masked attention, where each input frame is only allowed to attend to frames that precede it. We train the model to jointly predict the $6$D poses while also learning to accurately predict future features to match the true features. 
Such a setup has been employed in ~\cite{girdhar2021anticipative}, which shows that predicting future features is an effective self-supervised pretext task for learning visual representations. 

While the temporal architecture described above can be applied on top of any visual feature encoder (as discussed in ablations), we propose a purely transformer-based model that uses a Swin transformer~\cite{liu2022video} as the backbone. 
This enables our network to not only attend temporally to frames in the video, but also spatially within the frame.

% Motivated by this, in our proposed approach, we utilize a simple CNN-based architecture to extract useful features, and subsequently aggregate the information across consecutive frames using a recurrent neural network (RNN). 
% The training is performed on the YCB-Video dataset~\cite{xiang2018posecnn} and the approach achieves comparable performance to state-of-the-art approaches, while requiring lower computational costs.~We also conduct extensive ablation studies and demonstrate the effectiveness of our network design. 

In summary, the contributions of our paper are:

% Li et. al~\cite{li2018unified} use multiple frames from different viewing angles to estimate the poses, but do not use video frames with temporal consistency as the input. We propose to use video as input to perform refined pose estimation without using any downstream post-processing. We use convolutional neural networks (CNNs) to design a simple architecture to extract useful image features, followed by gated recurrent units (GRU) to aggregate information temporally across different frames. Without introducing additional computational burdens for feature extractions~\cite{xiang2018posecnn} or post-processing~\cite{deng2019poserbpf}, our simple network structure is able to achieve comparable performance on the challenging YCB dataset~\cite{xiang2018posecnn} with significantly improved computational speeds. 
\begin{itemize}
    \item We introduce a video based $6$D Object pose estimation framework that is purely attention based. 
    \item We incorporate self supervision via a predictive loss for learning better visual representations.
    \item We perform evaluation on  the challenging YCB-Video dataset~\cite{xiang2018posecnn}, where our algorithm achieves improved performance over state-of-the-art single frame methods such as PoseCNN~\cite{xiang2018posecnn} and PoseRBPF~\cite{deng2019poserbpf} with a real-time performance at 33fps, and transformer based method such as T6D-Dicrect~\cite{amini2021t6d}.
    \end{itemize}

\section{Related Work}
\label{sec:related_work}
% \irfan{This section is a bit verbose, but OK!} 
Estimating the $6$-DOF pose of objects in the scene is a widely studied task. The classical methods either use template-based or feature-based approaches. In template-based methods, a template is matched to different locations in the image, and a similarity score is computed~\cite{hinterstoisser2012model,hinterstoisser2011gradient}. However, these template matching methods could fail to make predictions for textureless objects and cluttered environments. In feature based methods, local features are extracted, and correspondence between known 3D objects and local 2D features is established using PnP to recover 6D poses~\cite{rothganger20063d,pavlakos20176}. However, these methods also require sufficient textures on the object to compute local features and face difficulty in generalising well to new environments as they are often trained on small datasets. 

% Classical methods rely on extracting local features from the input images, matching them with known $3$D objects, and subsequently running PnP to get $3$D-to-$2$D correspondences. However, extracting good features is key to obtaining good pose estimates. Therefore, over the years, much effort has been invested in designing good feature extractors and descriptors that are scale invariant and robust to noise and occlusions. Textureless objects were often handled using template matching methods \cite{hinterstoisser2012model,hinterstoisser2011gradient}. While template matching methods are effective for poorly textured objects, they fail in the presence of mild occlusions and cluttered backgrounds.

% review datasets
Convolutional Neural Networks (CNNs) have proven to be an effective tool in many computer vision tasks. However, they rely heavily on the availability of large-scale annotated datasets. Due to this limitation, the YCB-Video~\cite{xiang2018posecnn}, T-LESS~\cite{Hodan2017TLESSAR}, and OccludedLINEMOD datasets~\cite{krull2015learning,peng2019pvnet} were introduced. 
They have enabled the emergence of novel network designs such as PoseCNN~\cite{xiang2018posecnn}, DPOD~\cite{zakharov2019dpod}, PVNet~\cite{peng2019pvnet}, and others~\cite{bukschat2020efficientpose,wang2021gdr,he2021ffb6d,chen2022epro}. 
In this paper, we use the challenging YCB-Video dataset, as it is a popular dataset that serves as a testbed for many recent methods~\cite{amini2021t6d,amini2022yolopose,deng2019poserbpf}. 

Building on those datasets, various CNN architectures have been introduced to learn effective representations of objects and to estimate accurate 6D poses. Kehl~\etal~\cite{Kehl2017SSD6DMR} extend SSD~\cite{liu2016ssd} by adding an additional viewpoint classification branch to the network whereas ~\cite{Rad2017BB8AS,tekin2018real} predict 2D projections from 3D bounding box estimations. 
Other methods involve a hybrid approach where the model learns to perform multiple tasks, e.g., Song~\etal~\cite{song2020hybridpose} enforce consistencies among keypoints, edges, and object symmetries, and Billings~\etal~\cite{billings2019silhonet} predict silhouettes of objects along with object poses. 
There is also a growing trend of designing model agnostic features~\cite{wang2019normalized} that can handle novel objects. 
Finallly, few shot, one shot, and category level pose estimation has also seen increased interest recently~\cite{chen2021fs,wen2021bundletrack,sun2022onepose}
To refine the predicted poses, several works use additional depth information and perform the standard ICP algorithm~\cite{xiang2018posecnn,Kehl2017SSD6DMR}, directly learn from RGB-D inputs~\cite{wang2019densefusion,li2018deepim,zakharov2019dpod}, or through neural rendering~\cite{yen2021inerf,iwase2021repose,li2018deepim,ma2022robust}. 
We argue that since the input signals to robots and/or mobile devices are typically video sequences, instead of heavily relying on post processing refinement using additional depth information and rendering, estimating poses in videos by exploiting the temporal data could already refine the single pose estimations. 
Recently, several tracking algorithms are utilising videos to estimate object poses.
A notable work from Deng~\etal~\cite{deng2019poserbpf} introduces the PoseRBPF algorithm that uses particle filters to track objects in video sequences. 
However, this state-of-the-art algorithm provides accurate estimations at a high computational cost. 
Wen~\etal~\cite{wen2020se} also perform tracking, but use synthetic rendering of the object at the previous time-step. 

With the rise in self-attention models and Transformer architectures~\cite{vaswani2017attention,GPT2,yu2020ernie}, we also saw an increased interest in vision based transformers~\cite{arnab2021vivit,liu2022video,bao2021beit}. 
This has resulted in the application of Transformers to other applications like object detection~\cite{carion2020end,zhu2020deformable} and human pose estimation~\cite{zhao2022dpit,zheng20213d,li2022mhformer,panteleris2022pe,mao2021tfpose}, and object pose estimation~\cite{park2021dprost,amini2022yolopose,amini2021t6d}. 
TD6-Direct~\cite{amini2021t6d} builds on the Detection Transformer (DETR)~\cite{carion2020end} architecture to directly regress to the pose, while ~\cite{amini2022yolopose} uses transformers to predict keypoints, and subsequently does keypoint regression. 
In contrast to these works, we use Transformer models to attend over a set of frames in a video, and directly regress to the 6D pose. As transformers use a self-attention mechanism, our framework is capable of learning and refining $6$D poses from previous frames, without needing additional post process refinement.  
%\hspace{-5em}
\section{Approach}
\label{sec:methodology}

\begin{figure*}[]
    \centering
    \includegraphics[width=1\textwidth]{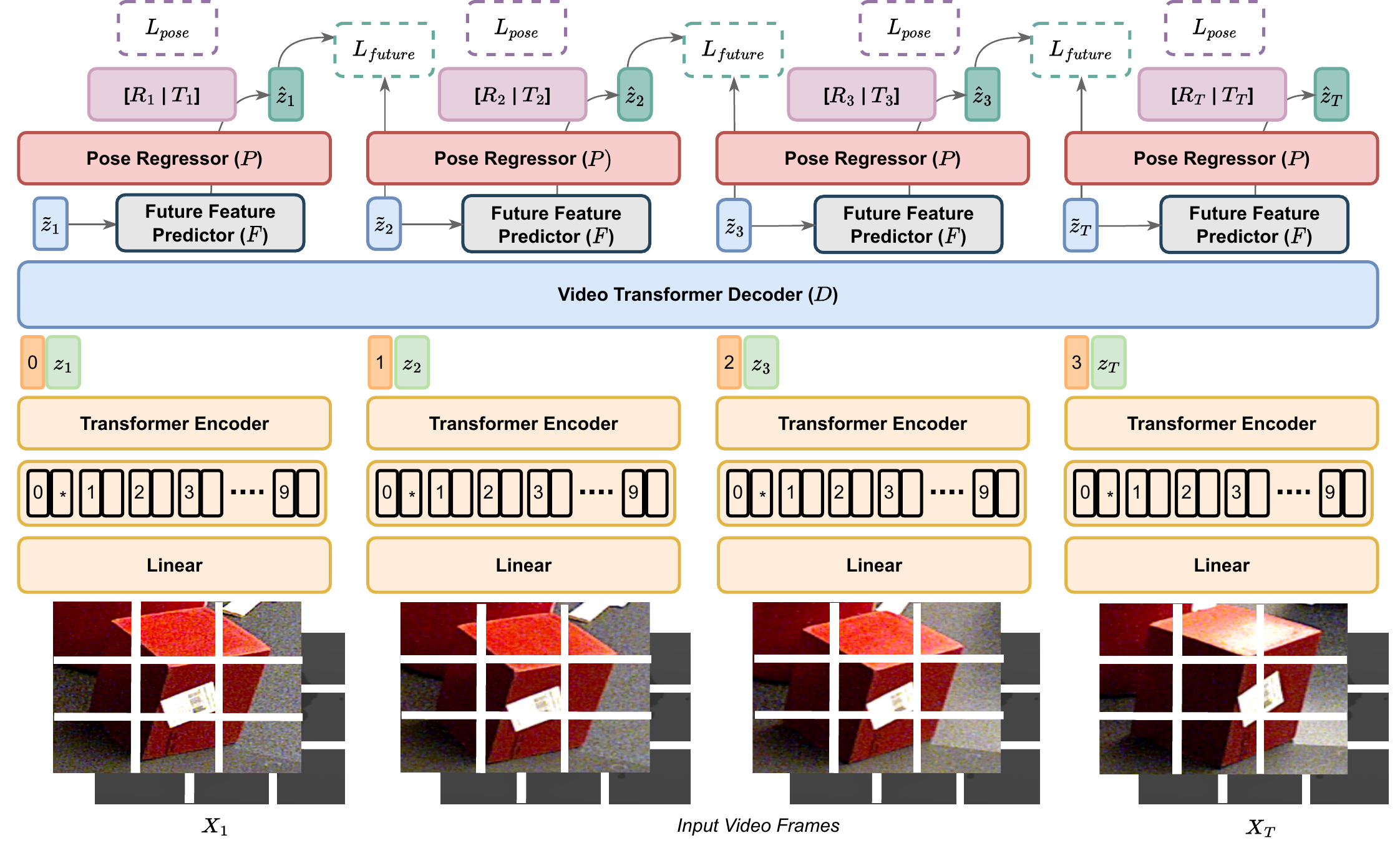} 
    \caption{Overview of our framework for 6D object pose estimation. We use Swin transformer~\cite{liu2022video} as Transformer Encoder, and GPT2~\cite{GPT2} as Video Transformer Decoder. Future Feature Predictor and Pose Regressor consists of a 2 layer MLP, further described in Figure~\ref{fig:regressors}}
    \label{fig:overview}
\end{figure*}
Given an RGB-D video stream, our goal is to estimate the $3$D rotation and $3$D translation of all the objects in every frame of the video. 
We assume the system has access to the $3$D model of the object. 
In the following sections, $\boldsymbol{R}$ denotes the rotation matrix with respect to the annotated canonical object pose, and $\boldsymbol{T}$ is the translation from the object to the camera.

\subsection{Overview of the network}
\label{sec: overview of the network}
Our pipeline, as shown in Figure \ref{fig:overview}, is a two stage network. The first stage comprises of a feature extraction module; We use a Swin transformer~\cite{liu2022video} to learn the visual features for every frame in the video. For a given video sequence and its corresponding depth, the transformer encoder gives us a feature vector of shape $b \times t  \times n  \times 768$ where $\textbf{b}$ corresponds to the batch size, $\textbf{t}$ corresponds to the temporal length and $\textbf{n}$ corresponds to the number of objects in the image. 

Pose Estimation relies on accurate object detection, which derives the class-id and Region-Of-Interest (ROI). 
During training, we use the ground truth bounding box whereas during testing, we use the predictions and bounding box from the PoseCNN model. 
This can potentially be replaced with any lightweight feature extraction model such as  MobileNet~\cite{howard2017mobilenets} to make the inference faster, or DETR~\cite{carion2020end} - a transformer based object detection module.
We also use depth as an additional input. 
In this paper, we use ground truth depth images; however, they can be replaced with other depth estimation modules.

%%%%%%%%%%%%%%%%%%%%%%%%%%%%%%%%%%%%%%%%%%%%%%%%%%%%%%%%%%%%%%%%%%%%%%%%%%%

\subsection{Video Transformer Decoder}
Given the features extracted by the encoder, the decoder attends to the previous features and predicts the 6D poses using the Pose Regressor, and the future frame features using Future Feature Predictor networks. 
We denote $\tilde{z}_T$ for the decoded features, and $\hat{z}_T$ for the predicted future features. 
Specifically, 
\begin{gather}
    \hat{z}_1, \cdots, \hat{z}_T = F(\tilde{z}_1, \cdots, \tilde{z}_T)
\end{gather}
And, 
\begin{gather}
    [R_T | T_T] = P(\tilde{z}_T)
\end{gather}

\subsection{6D Pose Regression}
\label{regression}
\begin{figure*}[]
    \centering
    \includegraphics[width=0.8\textwidth]{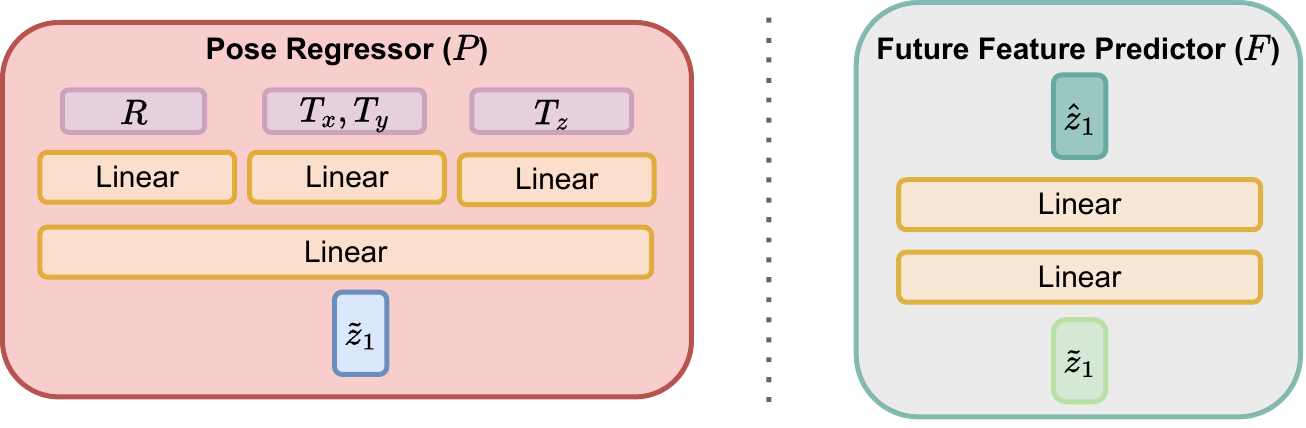} 
    \caption{(\textit{Left}) \textbf{Pose Regressor (P)}: Features from the temporal decoder is passed through a single linear layer, which is then passed through 3 separate linear layers for the estimation of $\boldsymbol{R}$, $\boldsymbol{(T_x, T_y)}$ and $\boldsymbol{T_z}$. (\textit{Right}) \textbf{Future Feature Predictor (F)}: Features from decoder are passed through a $2$ layer MLP to predict the future features.}
    \label{fig:regressors}
\end{figure*}

The Pose Regressor (\textit{P}) architecture can be seen in Figure~\ref{fig:regressors}.
Rotation $\boldsymbol{R}$ and Translation $\boldsymbol{T}$ have one common linear layer, which is then branched to three different linear layers. 
As $\boldsymbol{R}$, $\boldsymbol{(T_x, T_y)}$ and $\boldsymbol{T_z}$ occupy different latent spaces, we believe it is best to learn them separately, instead of a single vector of $7$ values as $\boldsymbol{[R | T]}$.
\subsubsection*{Translation}
The translation vector $\boldsymbol{T}$ is the object location in the camera coordinate system. A naive way of estimating $\boldsymbol{T}$ is to directly regress to it. However, doing so cannot handle multiple object instances or generalise well to new objects. To tackle this problem, Xiang~\etal~\cite{xiang2018posecnn} estimate $\boldsymbol{T}$ by localising the $2$D object center in the image and estimating object distance from the camera. Suppose \begin{math}\mathbf{c} = (c_x, c_y)^T \end{math}  are the centers of the object in the frame and $T_z$ is either learnt or estimated from the depth image, then \textit{$T_x$} and \textit{$T_y$} can be estimated as:
\begin{gather}
 \begin{bmatrix}c_x \\ c_y \end{bmatrix}
 =
 \begin{bmatrix}
  f_x\displaystyle\frac{T_x}{T_z}+ p_x \\
  f_y\displaystyle\frac{T_y}{T_z} + p_y
  \end{bmatrix}, \label{eq:1}
\end{gather} where \textit{$f_x$} and \textit{$f_y$} are focal lengths and $(\textit{$p_x$}, \textit{$p_y$})^\textit{T}$ are principal points. Since we have rough estimates of object locations from the noisy object detection inputs, we train our model to estimate \begin{math}\Delta c_x, \Delta c_y, \text{and} \ T_z \end{math}. We then estimate \textit{$T_x$} and \textit{$T_y$} using the following equation:
\begin{gather}
 \begin{bmatrix}c_x + \Delta c_x \\ c_y + \Delta c_y\end{bmatrix}
 =
 \begin{bmatrix}
  f_x\displaystyle\frac{T_x}{T_z}+ p_x \\
  f_y\displaystyle\frac{T_y}{T_z} + p_y
  \end{bmatrix}. \label{eq:2}
\end{gather}

\subsubsection*{Rotation}
Similar to \cite{xiang2018posecnn}, we represent the rotation $\boldsymbol{R}$ using quaternions: $\boldsymbol{R}=\{w, x, y, z\}$. However, we only predict the $x, y, z$ values, and infer $w$, the real value as: \begin{equation}
    w = 1 - norm(x, y ,z)
\end{equation}
We noticed that doing this helps the training process, and the quaternions learnt are more bounded.

% ##############################################################################
\subsection{Training Strategy}
\label{sec:training_stratergy}
The pose estimation loss is obtained by projecting the $3$D points using the estimated and ground truth pose, and then computing their $L_2$ distance:
\begin{equation}
L_\text{pose}(\mathbf{\widetilde{q}}, \mathbf{q}) = \frac{1}{m} \sum\limits_{x \in M} || (R(\mathbf{\widetilde{q}})x + \mathbf{\widetilde{t}}) - (R(\mathbf{q})x + \mathbf{t})|| ^2,
\end{equation} 
where $M$ denotes the set of 3D points and $m$ is total number of points. 
$R(\mathbf{\widetilde{q}})$ and $R(\mathbf{q})$ indicate the rotation matrix computed from the quaternion representation as in \cite{xiang2018posecnn}.
In addition, we also add a cosine loss on the quaternions, and a regularisation loss to force the norm of the quaternion to be 1. Quaternions that represent rotations are unit norm, and forcing the norm to be bounded by 1 helps in the learning process by eliminating all non-plausible vector combinations. 
\begin{equation}\label{eq:lreg}
    L_\text{reg} = ||1 - \text{norm}(\mathbf{\widetilde{q}}) ||,~~~       
    L_\text{inner\_prod} = 1 - \langle\mathbf{\widetilde{q}},\mathbf{q}\rangle.
\end{equation}
In addition to these losses, inspired by ~\cite{girdhar2021anticipative}, we add a future feature loss that is defined as:
\begin{gather}
    L_{future} = \sum\limits_{t=1}^{T-1} || \mathbf{\hat{z}_t} - \mathbf{\tilde{z}_{t+1}} ||_2^2
\end{gather}
The total loss is defined as 
\begin{equation} \label{eq:loss2}
    L(\mathbf{\widetilde{q}}, \mathbf{q}, \mathbf{\widetilde{t}}, \mathbf{t}, \mathbf{\hat{z}}, \mathbf{\tilde{z}}) = L_\text{future} + L_\text{pose} + L_\text{reg} + L_\text{inner\_prod}.
\end{equation} 

% ################################################################################
\subsection{Implementation}
\label{sec:implementation}
VideoPose is implemented using the PyTorch \cite{paszke2017automatic} framework. We use a learning rate of $1e^{-4}$ and the Adam optimiser~\cite{kingma2014adam} with a weight decay of $1e^{-6}$.
We use the ReduceOnPlateau scheduler that decreases the learning rate by a factor of 0.9 with a patience of 3 epochs.
We create video samples of 5 frames and train our model for 20 epochs with the learning schedule described above. The training was done on 4 A40 GPUs for 1 week or 20 epochs, whichever occurred first.

During training, we augment the input images with colour-jitter and noise, and the bounding box by extending the height and width randomly between $0$ and $10\%$ of the height and width of the object. 
While training the temporal block, we create videos with random time jumps in between. 
For instance, given a large video sequence, we create video samples $1:n:10*n$, where n is a random number between $1$ and $10$, thus forcing the model to account for small and large jumps between consecutive frames.

% %===============================================================================
\section{Experiments} 
\label{sec:experiments}
\begin{figure}[]
    \centering
    \includegraphics[width=0.8\textwidth]{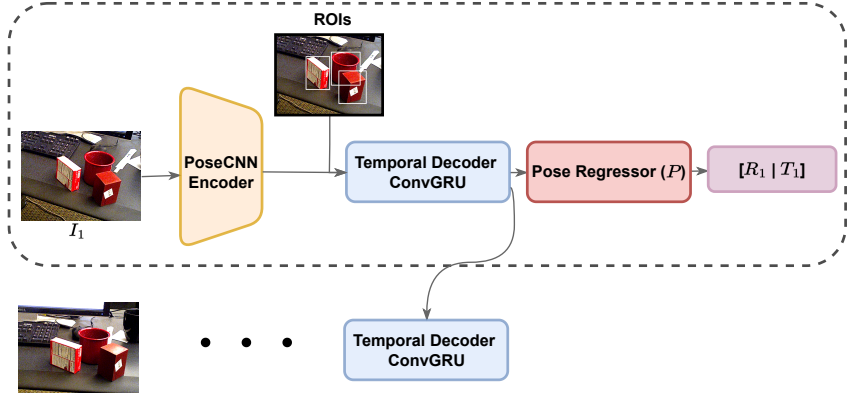} 
    \caption{Overview of our baseline framework for 6D object pose estimation. We use the PoseCNN encoder~\cite{xiang2018posecnn} as feature extractor, and a ConvGRU~\cite{ballas2015delving} as a temporal decoder. The regression module is similar to Figure~\ref{fig:regressors}}
    \label{fig:baseline}
\end{figure}

We compare our framework with PoseCNN\cite{xiang2018posecnn}, PoseRBPF\cite{deng2019poserbpf} and a recent image based transformer model T6D-Direct~\cite{amini2021t6d}. 
To the best of our knowledge, this work presents the first foray towards predicting pose directly from videos. 
Hence, we also create a simpler video baseline using the PoseCNN architecture for feature extraction, and ConvGRU to model the temporal information, as can be seen in Figure~\ref{fig:baseline}.
Further details about the baseline are provided in the Appendix~\ref{sec:appendix}.

\subsection{Dataset}
We evaluate the proposed method on the YCB-Video dataset \cite{xiang2018posecnn} (see Sec. \ref{sec:implementation} for reference). 
It contains 92 RGB-D video sequences of 21 objects, and contains both textured and textureless objects of varying shapes, and different levels of occlusion where about $15\%$ of objects are heavily occluded. 
Objects are annotated with 6D poses, segmentation masks and depth images. 

\subsection{Metrics}
We report the performance using two metrics: \emph{(i)} ADD, which is the average distance between the corresponding points of the 3D object at the ground truth and predicted poses; and, \emph{(ii)} ADD-S, which is designed for symmetric objects and calculates the mean distance from each $3$D point to a closest point on the target model. 
% We report the performance using two metrics: \emph{(i)} ADD, which is the average distance between the corresponding points of the 3D object at the ground truth and predicted poses; and, \emph{(ii)} ADD-S, designed for symmetric objects, . Given the estimated $[\mathbf{\widetilde{R}}|\mathbf{\widetilde{t}}]$ and the ground truth poses $[\mathbf{R}|\mathbf{t}]$, ADD-S, designed for symmetric objects, calculates the mean distance from each $3$D point to a closest point on the target model. 

\subsection{Evaluation}
\label{sec:result}
%%%%%%%%%%%%%%%%%%%%%%%%%%%%%%%%%%%%%%%%%%%%%%%%%%%%%%%%%%%%%%%%%%%%%%%%%%%

\begin{table}[]
\center
\caption{Comparison of performance between different architectures for single frame methods and video based baseline and our method, VideoPose. {\color[HTML]{009901} \textbf{Bold Green}} values represent the best method across the tracking/video methods (The corresponding methods are highlighted with {\color[gray]{0.5} gray columns}). \underline{Underlined} values compare between \textbf{ALL} the methods in the table.}
\setlength{\tabcolsep}{6pt} % Default value: 6pt
\renewcommand{\arraystretch}{1.2}
\resizebox{1\columnwidth}{!}{
\begin{tabular}{|c|cc|cc|
>{\columncolor[gray]{0.95}}c 
>{\columncolor[gray]{0.95}}c |
>{\columncolor[gray]{0.95}}c 
>{\columncolor[gray]{0.95}}c |
>{\columncolor[gray]{0.95}}c 
>{\columncolor[gray]{0.95}}c |
>{\columncolor[gray]{0.95}}c 
>{\columncolor[gray]{0.95}}c |}
\hline
 & \multicolumn{2}{c|}{PoseCNN} & \multicolumn{2}{c|}{DeepIM} & \multicolumn{2}{c|}{\cellcolor[gray]{0.95}\begin{tabular}[c]{@{}c@{}}PoseRBPF\\ (200 particles)\end{tabular}} & \multicolumn{2}{c|}{\cellcolor[gray]{0.95}\begin{tabular}[c]{@{}c@{}}PoseRBPF\\ (50 particles)\end{tabular}} & \multicolumn{2}{c|}{\cellcolor[gray]{0.95}\begin{tabular}[c]{@{}c@{}}VideoPose\\ Baseline\\ (ConvGRU)\end{tabular}} & \multicolumn{2}{c|}{\cellcolor[gray]{0.95}\begin{tabular}[c]{@{}c@{}}VideoPose \\ (Transformer)\end{tabular}} \\ \cline{2-13} 
 & \multicolumn{1}{c|}{ADD} & ADD-S & \multicolumn{1}{c|}{ADD} & ADD-S & \multicolumn{1}{c|}{\cellcolor[gray]{0.95}ADD} & ADD-S & \multicolumn{1}{c|}{\cellcolor[gray]{0.95}ADD} & ADD-S & \multicolumn{1}{c|}{\cellcolor[gray]{0.95}ADD} & ADD-S & \multicolumn{1}{c|}{\cellcolor[gray]{0.95}ADD} & ADD-S \\ \hline

002\_master\_chef\_can & 50.9 & 84.0 & \underline{ 71.2} & 93.1 & 58.0 & 77.1 & 56.1 & 75.6 & 36.3 & 80.9 & {\color[HTML]{009901} \textbf{70.2}} & {\color[HTML]{009901} \underline{ \textbf{93.3}}} \\
003\_cracker\_box & 51.7 & 76.9 & \underline{ 83.6} & \underline{ 91.0} & {\color[HTML]{009901} \textbf{76.8}} & {\color[HTML]{009901} \textbf{87.0}} & 73.4 & 85.2 & 22.5 & 62.1 & 43.3 & 78.2 \\
004\_sugar\_box & 68.6 & 84.3 & \underline{ 83.6} & \underline{ 91.0} & {\color[HTML]{009901} \textbf{75.9}} & {\color[HTML]{009901} \textbf{87.6}} & 73.9 & 86.5 & 40.7 & 68.8 & 58.1 & 82.5 \\
005\_tomato\_soup\_can & 66.0 & 80.9 & \underline{ 86.1} & 82.4 & 74.9 & 84.4 & 71.1 & 82.0 & 67.0 & 83.5 & {\color[HTML]{009901} \textbf{83.3}} & {\color[HTML]{009901} \underline{ \textbf{91.1}}} \\
006\_mustard\_bottle & 79.9 & 90.2 & \underline{ 91.5} & \underline{ 95.1} & {\color[HTML]{009901} \textbf{82.5}} & 91.0 & 80.0 & 90.0 & 75.3 & 89.0 & 75.2 & {\color[HTML]{009901} \textbf{91.8}} \\
007\_tuna\_fish\_can & 70.4 & 87.9 & \underline{ 87.7} & \underline{ 96.1} & 59.0 & 79.0 & 56.1 & 73.8 & 60.6 & 86.4 & {\color[HTML]{009901} \textbf{65.1}} & {\color[HTML]{009901} \textbf{94.0}} \\
008\_pudding\_box & 62.9 & 79.0 & \underline{ 82.7} & \underline{ 90.7} & 57.2 & 72.1 & 54.8 & 69.2 & 49.6 & 75.8 & {\color[HTML]{009901} \textbf{77.9}} & {\color[HTML]{009901} \textbf{90.3}} \\
009\_gelatin\_box & 75.2 & 87.1 & \underline{ 91.9} & \underline{ 94.3} & {\color[HTML]{009901} \textbf{88.8}} & 93.1 & 83.1 & 89.7 & 81.1 & 89.3 & 88.1 & {\color[HTML]{009901} \textbf{93.1}} \\
010\_potted\_meat\_can & 59.6 & 78.5 & \underline{ 76.2} & 86.4 & 49.3 & 62.0 & 47.0 & 61.3 & 61.5 & 83.6 & {\color[HTML]{009901} \textbf{71.1}} & {\color[HTML]{009901} \underline{ \textbf{89.3}}} \\
011\_banana & 72.3 & 85.9 & \underline{ 81.2} & \underline{ 91.3} & 24.8 & 61.5 & 22.8 & 64.2 & 22.3 & 69.6 & {\color[HTML]{009901} \textbf{54.3}} & {\color[HTML]{009901} \textbf{81.3}} \\
019\_pitcher\_base & 52.5 & 76.8 & \underline{ 90.1} & 94.6 & 75.3 & 88.4 & 74.0 & 87.5 & 70.5 & 85.9 & {\color[HTML]{009901} \textbf{78.0}} & {\color[HTML]{009901} \underline{ \textbf{90.6}}} \\
021\_bleach\_cleanser & 50.5 & 71.9 & \underline{ 81.2} & \underline{ 90.3} & 54.5 & 69.3 & 51.6 & 66.7 & 46.9 & 62.1 & {\color[HTML]{009901} \textbf{67.4}} & {\color[HTML]{009901} \textbf{88.4}} \\
024\_bowl & 6.5 & 69.7 & 8.6 & 81.4 & {\color[HTML]{009901} \underline{ \textbf{36.1}}} & 86.0 & 26.4 & {\color[HTML]{009901} \underline{ \textbf{88.2}}} & 12.8 & 80.1 & 13.0 & 78.8 \\
025\_mug & 57.7 & 78.0 & \underline{ 81.4} & 91.3 & {\color[HTML]{009901} \textbf{70.9}} & 85.4 & 67.3 & 83.7 & 67.6 & 88.8 & 54.1 & {\color[HTML]{009901} \underline{ \textbf{91.7}}} \\
035\_power\_drill & 55.1 & 75.8 & \underline{ 85.5} & \underline{ 92.3} & {\color[HTML]{009901} \textbf{70.9}} & {\color[HTML]{009901} \textbf{85.0}} & 64.4 & 80.6 & 36.0 & 71.2 & 58.8 & 82.7 \\
036\_wood\_block & 31.8 & 65.8 & \underline{ 60.0} & \underline{ 81.9} & 2.8 & 33.3 & 0.0 & 0.0 & 0.0 & 28.7 & {\color[HTML]{009901} \textbf{7.0}} & {\color[HTML]{009901} \textbf{68.6}} \\
037\_scissors & 35.8 & 56.2 & \underline{ 60.9} & \underline{ 75.4} & 21.7 & 33.0 & 20.6 & 30.9 & {\color[HTML]{009901} \textbf{50.1}} & {\color[HTML]{009901} \textbf{73.4}} & 21.2 & 60.8 \\
040\_large\_marker & 58.0 & 71.4 & \underline{ 75.6} & \underline{ 86.2} & 48.7 & 59.3 & 45.7 & 54.1 & 36.8 & 53.9 & {\color[HTML]{009901} \textbf{53.0}} & {\color[HTML]{009901} \textbf{84.2}} \\
051\_large\_clamp & 25.0 & 49.9 & \underline{ 48.4} & 74.3 & {\color[HTML]{009901} \textbf{47.3}} & 76.9 & 27.0 & 73.3 & 20.7 & 69.3 & 34.0 & {\color[HTML]{009901} \underline{ \textbf{81.8}}} \\
052\_extra\_large\_clamp & 15.8 & 47.0 & \underline{ 31.0} & \underline{ 73.3} & {\color[HTML]{009901} \textbf{26.5}} & {\color[HTML]{009901} \textbf{69.5}} & 50.4 & 68.7 & 5.9 & 55.4 & 8.0 & 60.6 \\
061\_foam\_brick & 40.4 & 87.8 & 35.9 & 81.9 & {\color[HTML]{009901} \underline{ \textbf{78.2}}} & {\color[HTML]{009901} \underline{ \textbf{89.7}}} & 75.8 & 88.4 & 44.8 & 86.1 & 45.7 & 92.7 \\ \hline
ALL & 53.7 & 75.9 & \underline{ 71.7} & \underline{ 88.1} & {\color[HTML]{009901} \textbf{59.9}} & 77.5 & 57.1 & 74.8 & 44.1 & 73.9 & 57.4 & {\color[HTML]{009901} \textbf{85.3}} \\ \hline
\end{tabular}
}
\vspace{0.1em}

\label{table:results_on_ycb}
\vspace{0.1em}
\end{table}

\begin{table}[]
\centering
\caption{Comparison of performance between transformer based object pose estimations and convolution based frameworks. {\color[HTML]{009901} \textbf{Bold Green}} values represent the best method across Transformer based methods (The corresponding methods are highlighted with {\color[gray]{0.5} gray columns}). \underline{Underlined} values compare between \textbf{ALL} the methods in the table.
T6D-Direct~\cite{amini2021t6d} and VideoPose Transformer frameworks are the Transformer based methods that have been compared and DeepIM, PoseRBPF and PoseCNN are CNN based methods.
{{\color[HTML]{3531FF} Blue} represents higher performance when bounding boxes from PoseCNN architecture is used.}
 }
\setlength{\tabcolsep}{6pt} % Default value: 6pt
\renewcommand{\arraystretch}{1.2}
\resizebox{1\columnwidth}{!}{
\begin{tabular}{|c|cc|cc|cc|
>{\columncolor[gray]{0.95}}c 
>{\columncolor[gray]{0.95}}c |
>{\columncolor[gray]{0.95}}c 
>{\columncolor[gray]{0.95}}c ||
>{\columncolor[gray]{0.95}}c 
>{\columncolor[gray]{0.95}}c |}
\hline
 & \multicolumn{2}{c|}{PoseCNN} & \multicolumn{2}{c|}{DeepIM} & \multicolumn{2}{c|}{\begin{tabular}[c]{@{}c@{}}PoseRBPF\\ (200 particles)\end{tabular}} & \multicolumn{2}{c|}{\cellcolor[gray]{0.95}T6D Direct} & \multicolumn{2}{c||}{\cellcolor[gray]{0.95}\begin{tabular}[c]{@{}c@{}}VideoPose \\ (Transformer)\end{tabular}} & \multicolumn{2}{c|}{\cellcolor[gray]{0.95}\begin{tabular}[c]{@{}c@{}}VideoPose \\ (Transformer)\\ (PoseCNN BBox\end{tabular}} \\ \cline{2-13} 
 & \multicolumn{1}{c|}{ADD} & ADD-S & \multicolumn{1}{c|}{ADD} & ADD-S & \multicolumn{1}{c|}{ADD} & ADD-S & \multicolumn{1}{c|}{\cellcolor[gray]{0.95}ADD} & ADD-S & \multicolumn{1}{c|}{\cellcolor[gray]{0.95}ADD} & ADD-S & \multicolumn{1}{c|}{\cellcolor[gray]{0.95}ADD} & \multicolumn{1}{c|}{\cellcolor[gray]{0.95}ADD-S} \\ \hline
002\_master\_chef\_can & 50.9 & 84.0 & \underline{ 71.2} & 93.1 & 58.0 & 77.1 & 61.5 & 91.9 & {\color[HTML]{009901} \textbf{70.2}} & {\color[HTML]{009901} \underline{ \textbf{93.3}}} & {\color[HTML]{3531FF} 68.7} & {\color[HTML]{3531FF} 92.0} \\
003\_cracker\_box & 51.7 & 76.9 & \underline{ 83.6} & \underline{ 91.0} & {\color[HTML]{333333} 76.8} & {\color[HTML]{333333} 87.0} & {\color[HTML]{009901} \textbf{76.3}} & {\color[HTML]{009901} \textbf{86.6}} & 43.3 & 78.2 & 31.6 & 75.1 \\
004\_sugar\_box & 68.6 & 84.3 & \underline{ 83.6} & \underline{ 91.0} & {\color[HTML]{333333} 75.9} & {\color[HTML]{333333} 87.6} & {\color[HTML]{009901} \textbf{81.8}} & {\color[HTML]{009901} \textbf{90.3}} & 58.1 & 82.5 & 52.2 & 80.7 \\
005\_tomato\_soup\_can & 66.0 & 80.9 & \underline{ 86.1} & 82.4 & 74.9 & 84.4 & 72.0 & 88.9 & {\color[HTML]{009901} \textbf{83.3}} & {\color[HTML]{009901} \underline{ \textbf{91.1}}} & {\color[HTML]{3531FF} 79.6} & {\color[HTML]{3531FF} 89.7} \\
006\_mustard\_bottle & 79.9 & 90.2 & \underline{ 91.5} & \underline{ 95.1} & {\color[HTML]{333333} 82.5} & {\color[HTML]{333333} 91.0} & {\color[HTML]{009901} \textbf{85.7}} & {\color[HTML]{009901} \textbf{94.7}} & 75.2 & {\color[HTML]{000000} 91.8} & 70.8 & 89.6 \\
007\_tuna\_fish\_can & 70.4 & 87.9 & \underline{ 87.7} & \underline{ 96.1} & 59.0 & 79.0 & 59.0 & 92.2 & {\color[HTML]{009901} \textbf{65.1}} & {\color[HTML]{009901} \textbf{94.0}} & {\color[HTML]{3531FF} 65.9} & {\color[HTML]{3531FF} 94.0} \\
008\_pudding\_box & 62.9 & 79.0 & \underline{ 82.7} & \underline{ 90.7} & 57.2 & 72.1 & 72.7 & 85.1 & {\color[HTML]{009901} \textbf{77.9}} & {\color[HTML]{009901} \textbf{90.3}} & 68.8 & {\color[HTML]{3531FF} 87.9} \\
009\_gelatin\_box & 75.2 & 87.1 & \underline{ 91.9} & \underline{ 94.3} & {\color[HTML]{333333} 88.8} & {\color[HTML]{333333} 93.1} & 74.4 & 86.9 & {\color[HTML]{009901} \textbf{88.1}} & {\color[HTML]{009901} \textbf{93.1}} & {\color[HTML]{3531FF} 87.1} & {\color[HTML]{3531FF} 93.0} \\
010\_potted\_meat\_can & 59.6 & 78.5 & \underline{ 76.2} & 86.4 & 49.3 & 62.0 & 67.8 & 83.5 & {\color[HTML]{009901} \textbf{71.1}} & {\color[HTML]{009901} \underline{ \textbf{89.3}}} & {\color[HTML]{3531FF} 70.2} & {\color[HTML]{3531FF} 88.4} \\
011\_banana & 72.3 & 85.9 & 81.2 & 91.3 & 24.8 & 61.5 & {\color[HTML]{009901} \underline{ \textbf{87.4}}} & {\color[HTML]{009901} \underline{ \textbf{93.8}}} & {\color[HTML]{333333} 54.3} & {\color[HTML]{333333} 81.3} & 50.6 & 81.5 \\
019\_pitcher\_base & 52.5 & 76.8 & \underline{ 90.1} & \underline{ 94.6} & {\color[HTML]{333333} 75.3} & {\color[HTML]{333333} 88.4} & {\color[HTML]{009901} \textbf{84.5}} & {\color[HTML]{009901} \textbf{92.3}} & 78.0 & 90.6 & 73.0 & 88.8 \\
021\_bleach\_cleanser & 50.5 & 71.9 & \underline{ 81.2} & \underline{ 90.3} & 54.5 & 69.3 & 65.0 & 83.0 & {\color[HTML]{009901} \textbf{67.4}} & {\color[HTML]{009901} \textbf{88.4}} & 56.0 & 81.3 \\
024\_bowl & 6.5 & 69.7 & 81.4 & 81.4 & {\color[HTML]{333333} 36.1} & {\color[HTML]{333333} 86.0} & {\color[HTML]{009901} \underline{ \textbf{91.6}}} & {\color[HTML]{009901} \underline{ \textbf{91.6}}} & 78.8 & 78.8 & 77.8 & 77.8 \\
025\_mug & 57.7 & 78.0 & \underline{ 81.4} & 91.3 & {\color[HTML]{333333} 70.9} & 85.4 & {\color[HTML]{009901} \textbf{72.1}} & 89.8 & 54.1 & {\color[HTML]{009901} \underline{ \textbf{91.7}}} & 48.2 & {\color[HTML]{3531FF} 91.5} \\
035\_power\_drill & 55.1 & 75.8 & \underline{ 85.5} & \underline{ 92.3} & {\color[HTML]{333333} 70.9} & {\color[HTML]{333333} 85.0} & {\color[HTML]{009901} \textbf{77.7}} & {\color[HTML]{009901} \textbf{88.8}} & 58.8 & 82.7 & 49.8 & 81.7 \\
036\_wood\_block & 31.8 & 65.8 & 81.9 & 81.9 & 2.8 & 33.3 & {\color[HTML]{009901} \underline{ \textbf{90.7}}} & {\color[HTML]{009901} \underline{ \textbf{90.7}}} & {\color[HTML]{333333} 68.6} & {\color[HTML]{333333} 68.6} & 66.4 & 66.4 \\
037\_scissors & 35.8 & 56.2 & \underline{ 60.9} & \underline{ 75.4} & 21.7 & 33.0 & {\color[HTML]{009901} \textbf{59.7}} & {\color[HTML]{009901} \textbf{83.0}} & {\color[HTML]{333333} 21.2} & {\color[HTML]{333333} 60.8} & 43.6 & 74.3 \\
040\_large\_marker & 58.0 & 71.4 & \underline{ 75.6} & \underline{ 86.2} & {\color[HTML]{333333} 48.7} & 59.3 & {\color[HTML]{009901} \textbf{63.9}} & 74.9 & 53.0 & {\color[HTML]{009901} \textbf{84.2}} & 51.5 & {\color[HTML]{3531FF} 82.1} \\
051\_large\_clamp & 25.0 & 49.9 & 74.3 & 74.3 & {\color[HTML]{333333} 47.3} & 76.9 & 78.3 & 78.3 & {\color[HTML]{009901} \underline{ \textbf{81.8}}} & {\color[HTML]{009901} \underline{ \textbf{81.8}}} & 67.9 & 67.9 \\
052\_extra\_large\_clamp & 15.8 & 47.0 & \underline{ 73.3} & \underline{ 73.3} & {\color[HTML]{333333} 26.5} & {\color[HTML]{333333} 69.5} & 54.7 & 54.7 & {\color[HTML]{009901} \textbf{60.6}} & {\color[HTML]{009901} \textbf{60.6}} & 54.4 & 54.4 \\
061\_foam\_brick & 40.4 & 87.8 & 81.9 & 81.9 & {\color[HTML]{333333} 78.2} & {\color[HTML]{333333} 89.7} & 89.9 & 89.9 & {\color[HTML]{009901} \underline{ \textbf{92.7}}} & {\color[HTML]{009901} \underline{ \textbf{92.7}}} & {\color[HTML]{3531FF} 92.7} & {\color[HTML]{3531FF} 92.7} \\ \hline
ALL & 53.7 & 75.9 & 71.7 & \underline{ 88.1} & {\color[HTML]{333333} 59.9} & 77.5 & {\color[HTML]{009901} \underline{ \textbf{74.6}}} & {\color[HTML]{009901} \textbf{86.2}} & 66.4 & {\color[HTML]{000000} 85.3} & 61.5 & 82.9 \\ \hline
\end{tabular}
}
\vspace{0.2em}
\label{table:results_on_transformer}
\vspace{0.1em}
\end{table}

We compare our results with PoseCNN \cite{xiang2018posecnn} for single frame prediction and PoseRBPF \cite{deng2019poserbpf} for videos in Table \ref{table:results_on_ycb}. We also compare our results against DeepIM~\cite{li2018deepim}. However, it is worth noting that the DeepIM is an iterative refinement framework that works on top of the PoseCNN results.
PoseRBPF does not directly utilise video information, but uses a tracking pipeline to iteratively refine their poses. 
Several newer image based pose estimation methods have redefined the state-of-the-art, but for the scope of this paper, and for cleaner comparisons, we compare our framework with the aforementioned works.
To the best of our knowledge, we believe that our work is the first to use videos directly in the estimation of $6$D Object Pose. 
Therefore, we also compare against a simple CNN and ConvGRU based framework as shown in Figure ~\ref{fig:baseline}. 

\begin{figure*}[]
\centering
\centering
    \includegraphics[width=0.9\textwidth]{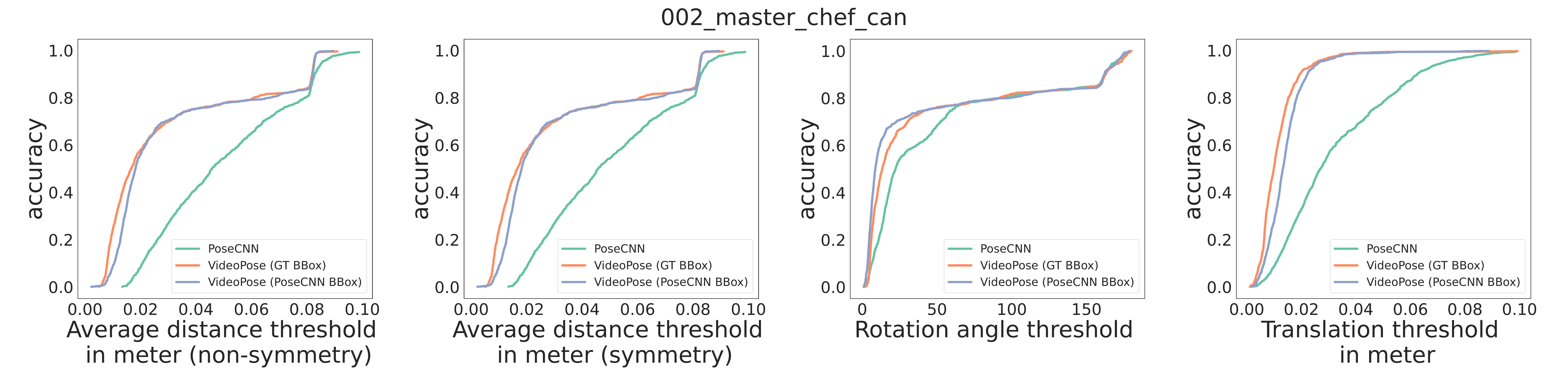}
    \includegraphics[width=0.9\textwidth]{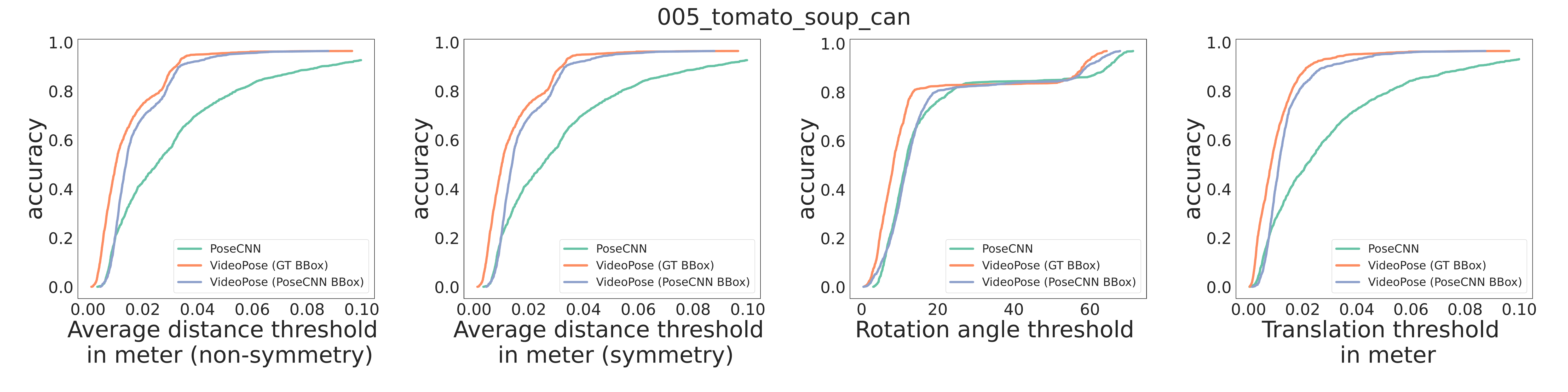}
    \includegraphics[width=0.9\textwidth]{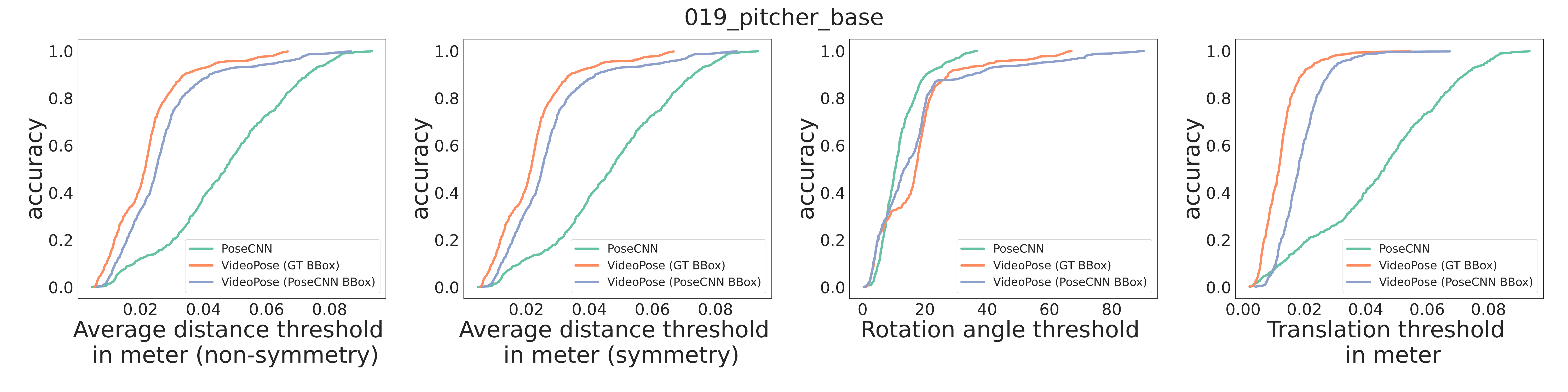}
    \includegraphics[width=0.9\textwidth]{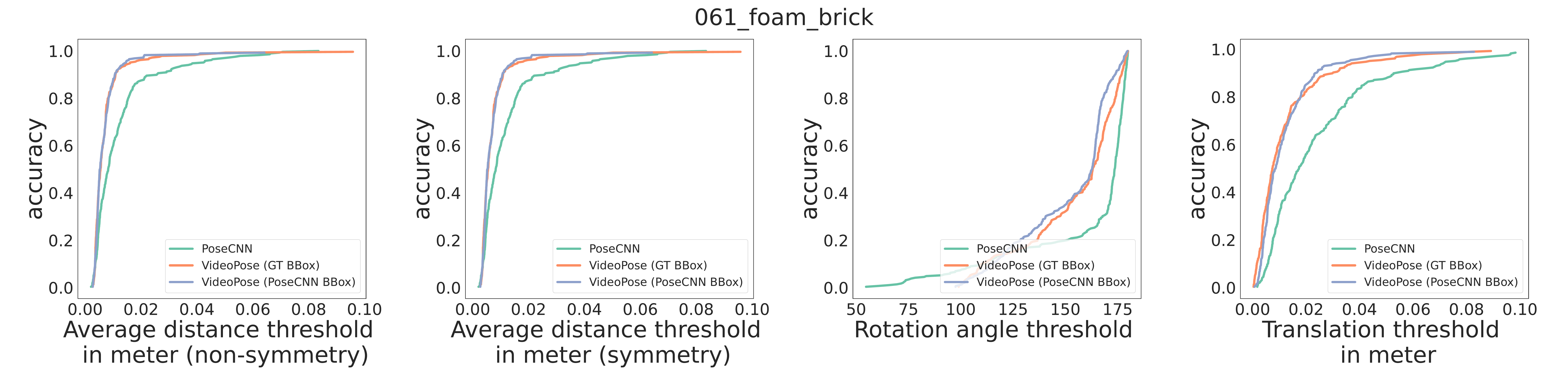}

\caption{Accuracy-threshold curve for rotation and translation, and reprojection errors for few objects in the YCB-Video dataset. \textcolor{Emerald}{Emerald} is used to plot PoseCNN curve, \textcolor{Peach}{Orange} for VideoPose Transformer with GT bounding boxes, and \textcolor{Orchid}{Purple} for VideoPose Transformer with bounding boxes from PoseCNN architectures. }
\label{fig:auc}
\end{figure*}

From Table ~\ref{table:results_on_ycb}, we see that when compared to PoseRBPF, the simpler VideoPose (with the ConvGRU) shows improvements for several objects (e.g 037\_scissors, 007\_tuna\_fish\_can, 010\_potted\_mean\_can etc.), whereas the Transformer based VideoPose outperforms PoseRBPF for most objects in ADD metric, and significantly outperforms for the ADD-S metric. 
These results indicate the effectiveness of using videos, and also the effectiveness of using attention in learning to estimate object poses.
In Table ~\ref{table:results_on_transformer}, we compare T6D-Direct against our method, and we see that our framework is able to perform better for most of the objects, and comparably to the rest.
VideoPose outperforms T6D-Direct for 12 of the 21 (57\%) objects for the ADD-S metric, and 10 of the 21 objects for ADD metric. 
T6D-Direct, in addition to pose, also predicts the bounding boxes and class probabilities. We believe that these additional auxiliary losses have aided the model's performance, and as a future work, we would like to explore this space.
In Table ~\ref{table:results_on_transformer}, we also provide results from using PoseCNN bounding boxes as opposed to GT bounding boxes, and we see that the performance drop is very minimal suggesting that the augmentation during the training has helped in the model learning to fix these predictions.
AUC in Figure ~\ref{fig:auc} shows that even for a difficult object such as a foam brick, our model performs significantly better than PoseCNN, and when the bounding boxes are not accurate, our method shows very slight drop in performance. 

% \noindent\textbf{Impact of different temporal blocks} We also perform ablation studies on different architectures used to capture the temporality, as shown in Table \ref{table:diff_arch}. 
% Instead of the baseline temporal RNN in fig. ~\ref{fig:simpleRNN}, we use a ConvGRU ~\cite{ballas2015delving} as the temporal module and observe that it handles the temporal information more effectively. The performance is comparable to or better than PoseCNN and PoseRBPF for more than 50\% of the objects. 
% We treat this ablation study as a proof that using previous estimates can aid in the pose estimation, regardless of the temporal module used.

\noindent\textbf{Effect of the number of previous frames used} Table \ref{table:comparison_video_length} shows the effect of number of previous frames used. We see that we get the best performance when $10$ frames are used, indicating that the model performance is influenced by previous frames. It is worth noting that the model is trained for a video length of $5$ frames. However, the model is still capable of extracting relevant information for longer-range videos. We believe that the augmentation described in Section ~\ref{sec:implementation} has helped the model in learning the long-range dependencies. 

\noindent\textbf{Time efficiency} We compare the time efficiency of our model with several models. 
The run time for PoseCNN is taken from Wang \etal~\cite{wang2019densefusion}. 
From Table \ref{table:time_efficiency}, we see that VideoPose with the Beit Transformer~\cite{bao2021beit} backbone performs the fastest, as opposed to the Swin Transformer used in our framework results. 
% It is worth noting that all the accuracies reported in this paper for VideoPose Transformer method uses Swin Transformer as the backbone feature extractor. 
We saw a significant improvement in accuracy when using Swin Transformer compared to Beit Transformer, at a small decrease in the speed, and hence reported results for Swin backbone throughout the paper. 
However, this also indicates that the our framework's speed is affected by the backbone framework, and learning from video adds a very small time cost. In the future, when faster, and more accurate attention based image feature extractors are developed, our framework can be easily adapted to it.
Our model is tested on GeForce GTX 1080 Ti. \\

\begin{table}[]
\begin{minipage}[c]{0.495\textwidth}
\caption{Comparison of frame rates for different methods}
\centering
\setlength{\tabcolsep}{8pt} % Default value: 6pt
\renewcommand{\arraystretch}{1.2}
\resizebox{0.7\columnwidth}{!}{
\begin{tabular}{|c|c|}
\hline
Method & Time (fps) \\ \hline
PoseCNN~\cite{xiang2018posecnn} & 5.88 \\ \hline
PoseRBPF(50)~\cite{deng2019poserbpf} & 20 \\ \hline
PoseRBPF(200)~\cite{deng2019poserbpf} & 5 \\ \hline
CosyPose~\cite{labbe2020cosypose} & 3 \\ \hline
DeepIM ~\cite{li2018deepim} & 12 \\ \hline
T6D-Direct~\cite{amini2021t6d} & 59 \\ \hline
Ours(Swin backbone) & 33 \\ \hline
Ours(Beit backbone) & \textbf{67}  \\ \hline  
\end{tabular}
}
\vspace{.05in} 
\label{table:time_efficiency}
\end{minipage}
\hspace{0.1in}
\begin{minipage}[c]{0.4\textwidth}
\vspace{-0.4in} 
\caption{Effect of video length on accuracies.}
\vspace{0.3in} 
\centering
\setlength{\tabcolsep}{8pt} % Default value: 6pt
\renewcommand{\arraystretch}{1.2}
\resizebox{0.7\columnwidth}{!}{
\begin{tabular}{|c|c|c|}
\hline
Video Length & ADD & ADD-S \\ \hline
2 & 63.9 & 84 \\ \hline
5 & 64.3 & 84.4 \\ \hline
7 & 64.1 & 84.2 \\ \hline
10 & \textbf{64.8} & \textbf{84.8} \\ \hline
\end{tabular}
}
\vspace{0.in} 

\label{table:comparison_video_length}
\vspace{.1in} 
\end{minipage}
\centering
\end{table}

\begin{figure}[!t]
    \centering
    \includegraphics[width=0.9\textwidth]{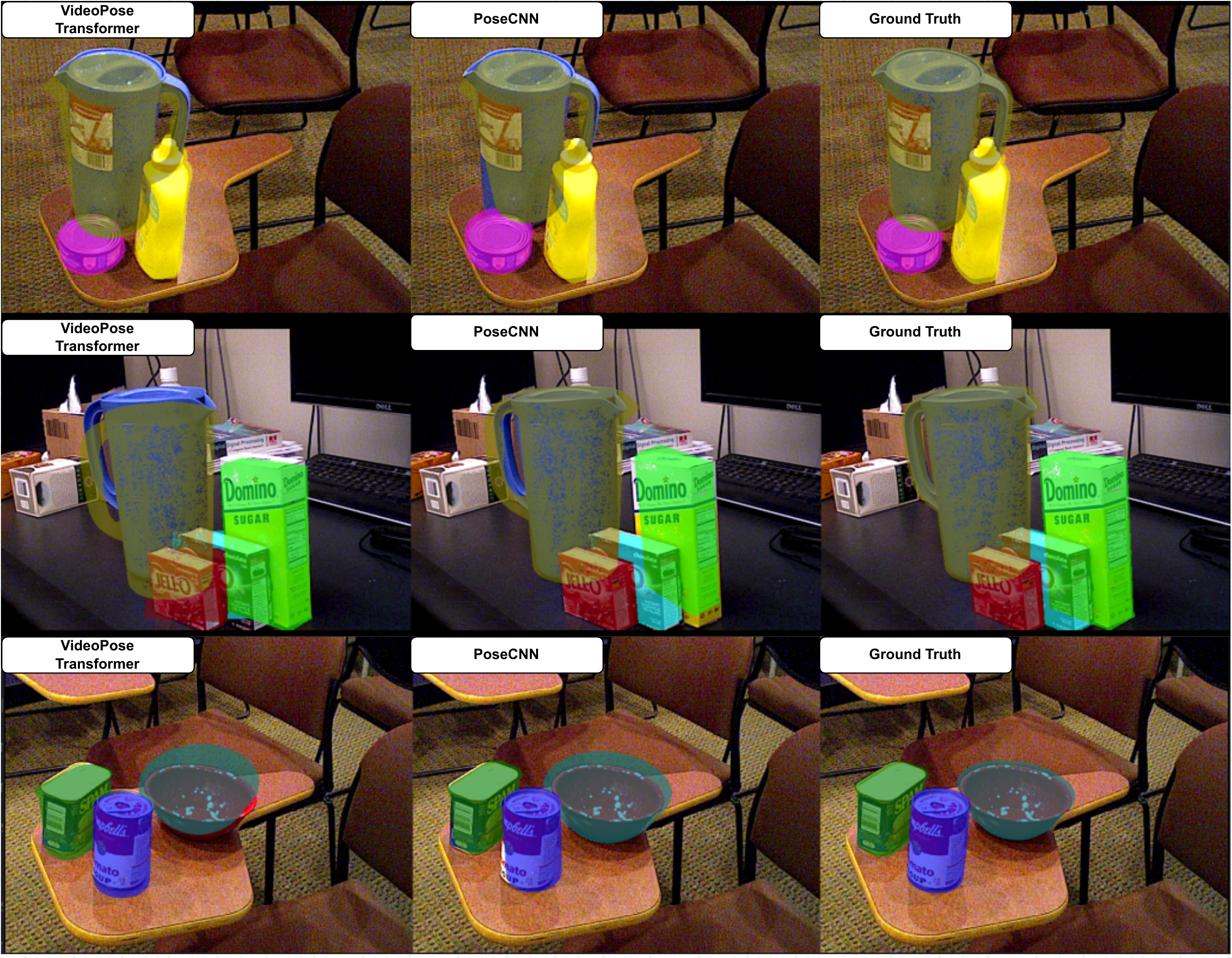} 
    \caption{Visualisations of the estimated poses on YCB-Dataset: each row represents results at different time-steps. The columns consist of VideoPose, PoseCNN and Ground truth visualisations respectively.}
    \label{fig:qualitative1}
\end{figure}

\noindent \textbf{Qualitative Analysis of the $6$D predictions} We show three examples of the predictions by VideoPose Transformer, PoseCNN, and ground truth poses in Table \ref{fig:qualitative1}. The columns represent the $2$D projections of predictions using VideoPose Transformer, PoseCNN, and the ground truth poses. 
We observe that the poses estimated using the VideoPose Transformer framework looks qualitatively more accurate than PoseCNN.\\

\section{Conclusion}
\label{sec:conclusion}
In this work, we introduced an end-to-end self-attention based Transformer network for the estimation of $6$D Object Pose called \textit{VideoPose}.
We demonstrate that by using the $6$D predictions from the previous frames, we can significantly improve $6$D predictions in the subsequent frames. 
We also conducted an extensive ablation study on different design choices of the network, and show that our model is able to learn and utilise the features from previous predictions regardless of the network choices. 
Finally, the proposed network performs in real-time at 33fps, thereby improving the time efficiency over previous approaches. 
As a future work, we would like to further improve our architecture with a better temporal module and model the relationship with the camera transformation and the objects. 
As T6D-Direct showed improvement with a single-stage network, detecting objects along with their poses, we would like to explore this idea in our future work as well.
Our method successfully maintains consistency in pose estimation between frames, however, still depends on the initial frame estimation, and accurate bounding box prediction. 
We would like to investigate further on improving this, while maintaining/improving the computational efficiency.

% In this work, we introduce VideoPose, a simple convolutional neural network architecture to estimate object 6D poses from videos. We demonstrate that by using the $6$D predictions from the previous frames, we can significantly improve $6$D predictions in the subsequent frames. We also conduct an extensive ablation study on different design choices of the network, and show that our model is able to learn and utilise the features from previous predictions regardless of the network choices. Finally, the proposed network performs in real-time at 30fps, thereby improving the time efficiency over previous approaches. As a future work, we would like to further improve our architecture with a better temporal module and model the relationship with the camera transformation and the objects. Our method successfully maintains consistency in pose estimation between frames, however, still depends on the initial frame estimation. We would like to investigate further on improving this, while maintaining the computational efficiency. \\
%===============================================================================

\clearpage
{
\small
\bibliographystyle{ieee_fullname}
\bibliography{egbib}

\begin{thebibliography}{10}\itemsep=-1pt

\bibitem{amini2021t6d}
Arash Amini, Arul~Selvam Periyasamy, and Sven Behnke.
\newblock T6d-direct: Transformers for multi-object 6d pose direct regression.
\newblock In {\em DAGM German Conference on Pattern Recognition}, pages
  530--544. Springer, 2021.

\bibitem{amini2022yolopose}
Arash Amini, Arul~Selvam Periyasamy, and Sven Behnke.
\newblock Yolopose: Transformer-based multi-object 6d pose estimation using
  keypoint regression.
\newblock {\em arXiv preprint arXiv:2205.02536}, 2022.

\bibitem{arnab2021vivit}
Anurag Arnab, Mostafa Dehghani, Georg Heigold, Chen Sun, Mario Lu{\v{c}}i{\'c},
  and Cordelia Schmid.
\newblock Vivit: A video vision transformer.
\newblock In {\em Proceedings of the IEEE/CVF International Conference on
  Computer Vision}, pages 6836--6846, 2021.

\bibitem{ballas2015delving}
Nicolas Ballas, Li Yao, Chris Pal, and Aaron Courville.
\newblock Delving deeper into convolutional networks for learning video
  representations.
\newblock {\em arXiv preprint arXiv:1511.06432}, 2015.

\bibitem{bao2021beit}
Hangbo Bao, Li Dong, and Furu Wei.
\newblock Beit: Bert pre-training of image transformers.
\newblock {\em arXiv preprint arXiv:2106.08254}, 2021.

\bibitem{billings2019silhonet}
Gideon Billings and Matthew Johnson-Roberson.
\newblock Silhonet: An rgb method for 6d object pose estimation.
\newblock {\em IEEE Robotics and Automation Letters}, 4(4):3727--3734, 2019.

\bibitem{Brachmann2014Learning}
Eric Brachmann, Alexander Krull, Frank Michel, Stefan Gumhold, Jamie Shotton,
  and Carsten Rother.
\newblock Learning {6D} object pose estimation using 3d object coordinates.
\newblock In {\em European conference on computer vision}, pages 536--551.
  Springer, 2014.

\bibitem{bukschat2020efficientpose}
Yannick Bukschat and Marcus Vetter.
\newblock Efficientpose: An efficient, accurate and scalable end-to-end 6d
  multi object pose estimation approach.
\newblock {\em arXiv preprint arXiv:2011.04307}, 2020.

\bibitem{carion2020end}
Nicolas Carion, Francisco Massa, Gabriel Synnaeve, Nicolas Usunier, Alexander
  Kirillov, and Sergey Zagoruyko.
\newblock End-to-end object detection with transformers.
\newblock In {\em European conference on computer vision}, pages 213--229.
  Springer, 2020.

\bibitem{chen2022epro}
Hansheng Chen, Pichao Wang, Fan Wang, Wei Tian, Lu Xiong, and Hao Li.
\newblock Epro-pnp: Generalized end-to-end probabilistic perspective-n-points
  for monocular object pose estimation.
\newblock In {\em Proceedings of the IEEE/CVF Conference on Computer Vision and
  Pattern Recognition}, pages 2781--2790, 2022.

\bibitem{chen2021fs}
Wei Chen, Xi Jia, Hyung~Jin Chang, Jinming Duan, Linlin Shen, and Ales
  Leonardis.
\newblock Fs-net: Fast shape-based network for category-level 6d object pose
  estimation with decoupled rotation mechanism.
\newblock In {\em Proceedings of the IEEE/CVF Conference on Computer Vision and
  Pattern Recognition}, pages 1581--1590, 2021.

\bibitem{chen2017multi}
Xiaozhi Chen, Huimin Ma, Ji Wan, Bo Li, and Tian Xia.
\newblock Multi-view {3D} object detection network for autonomous driving.
\newblock In {\em Proceedings of the IEEE Conference on Computer Vision and
  Pattern Recognition (CVPR)}, pages 1907--1915, 2017.

\bibitem{deng2019poserbpf}
Xinke Deng, Arsalan Mousavian, Yu Xiang, Fei Xia, Timothy Bretl, and Dieter
  Fox.
\newblock Poserbpf: A rao-blackwellized particle filter for 6d object pose
  tracking.
\newblock {\em Robotics: Science and Systems (RSS)}, 2019.

\bibitem{eppner2019billion}
Clemens Eppner, Arsalan Mousavian, and Dieter Fox.
\newblock A billion ways to grasp: An evaluation of grasp sampling schemes on a
  dense, physics-based grasp data set.
\newblock {\em arXiv preprint arXiv:1912.05604}, 2019.

\bibitem{girdhar2021anticipative}
Rohit Girdhar and Kristen Grauman.
\newblock {Anticipative Video Transformer}.
\newblock In {\em ICCV}, 2021.

\bibitem{he2021ffb6d}
Yisheng He, Haibin Huang, Haoqiang Fan, Qifeng Chen, and Jian Sun.
\newblock Ffb6d: A full flow bidirectional fusion network for 6d pose
  estimation.
\newblock In {\em Proceedings of the IEEE/CVF Conference on Computer Vision and
  Pattern Recognition}, pages 3003--3013, 2021.

\bibitem{hinterstoisser2011gradient}
Stefan Hinterstoisser, Cedric Cagniart, Slobodan Ilic, Peter Sturm, Nassir
  Navab, Pascal Fua, and Vincent Lepetit.
\newblock Gradient response maps for real-time detection of textureless
  objects.
\newblock {\em IEEE Transactions on Pattern Analysis and Machine Intelligence
  (TPAMI)}, 34(5):876--888, 2011.

\bibitem{hinterstoisser2012model}
Stefan Hinterstoisser, Vincent Lepetit, Slobodan Ilic, Stefan Holzer, Gary
  Bradski, Kurt Konolige, and Nassir Navab.
\newblock Model based training, detection and pose estimation of texture-less
  3d objects in heavily cluttered scenes.
\newblock In {\em Proceedings of the Asian Conference on Computer Vision
  (ACCV)}, pages 548--562. Springer, 2012.

\bibitem{Hodan2017TLESSAR}
Tom{\'a}s Hodan, Pavel Haluza, Step{\'a}n Obdrz{\'a}lek, Jiri Matas, Manolis
  I.~A. Lourakis, and Xenophon Zabulis.
\newblock T-less: An rgb-d dataset for 6d pose estimation of texture-less
  objects.
\newblock {\em IEEE Winter Conference on Applications of Computer Vision
  (WACV)}, pages 880--888, 2017.

\bibitem{howard2017mobilenets}
Andrew~G Howard, Menglong Zhu, Bo Chen, Dmitry Kalenichenko, Weijun Wang,
  Tobias Weyand, Marco Andreetto, and Hartwig Adam.
\newblock {MobileNets}: Efficient convolutional neural networks for mobile
  vision applications.
\newblock {\em arXiv preprint arXiv:1704.04861}, 2017.

\bibitem{hu2019segmentation}
Yinlin Hu, Joachim Hugonot, Pascal Fua, and Mathieu Salzmann.
\newblock Segmentation-driven 6d object pose estimation.
\newblock In {\em Proceedings of the IEEE Conference on Computer Vision and
  Pattern Recognition (CVPR)}, pages 3385--3394, 2019.

\bibitem{iwase2021repose}
Shun Iwase, Xingyu Liu, Rawal Khirodkar, Rio Yokota, and Kris~M Kitani.
\newblock Repose: Fast 6d object pose refinement via deep texture rendering.
\newblock In {\em Proceedings of the IEEE/CVF International Conference on
  Computer Vision}, pages 3303--3312, 2021.

\bibitem{Kehl2017SSD6DMR}
Wadim Kehl, Fabian Manhardt, Federico Tombari, Slobodan Ilic, and Nassir Navab.
\newblock Ssd-6d: Making rgb-based 3d detection and 6d pose estimation great
  again.
\newblock {\em Proc. of the IEEE International Conference on Computer Vision
  (ICCV)}, pages 1530--1538, 2017.

\bibitem{kingma2014adam}
Diederik~P Kingma and Jimmy Ba.
\newblock Adam: A method for stochastic optimization.
\newblock {\em arXiv preprint arXiv:1412.6980}, 2014.

\bibitem{konishi2018real}
Yoshinori Konishi, Kosuke Hattori, and Manabu Hashimoto.
\newblock Real-time {6D} object pose estimation on {CPU}.
\newblock {\em arXiv preprint arXiv:1811.08588}, 2018.

\bibitem{krull2015learning}
Alexander Krull, Eric Brachmann, Frank Michel, Michael Ying~Yang, Stefan
  Gumhold, and Carsten Rother.
\newblock Learning analysis-by-synthesis for 6d pose estimation in rgb-d
  images.
\newblock In {\em Proc. of the IEEE International Conference on Computer Vision
  (ICCV)}, pages 954--962, 2015.

\bibitem{labbe2020cosypose}
Yann Labb{\'e}, Justin Carpentier, Mathieu Aubry, and Josef Sivic.
\newblock Cosypose: Consistent multi-view multi-object 6d pose estimation.
\newblock In {\em European Conference on Computer Vision}, pages 574--591.
  Springer, 2020.

\bibitem{li2018unified}
Chi Li, Jin Bai, and Gregory~D Hager.
\newblock A unified framework for multi-view multi-class object pose
  estimation.
\newblock In {\em Proceedings of the European Conference on Computer Vision
  (ECCV)}, pages 254--269, 2018.

\bibitem{li2022mhformer}
Wenhao Li, Hong Liu, Hao Tang, Pichao Wang, and Luc Van~Gool.
\newblock Mhformer: Multi-hypothesis transformer for 3d human pose estimation.
\newblock In {\em Proceedings of the IEEE/CVF Conference on Computer Vision and
  Pattern Recognition}, pages 13147--13156, 2022.

\bibitem{li2018deepim}
Yi Li, Gu Wang, Xiangyang Ji, Yu Xiang, and Dieter Fox.
\newblock {DeepIm}: Deep iterative matching for 6d pose estimation.
\newblock In {\em Proceedings of the European Conference on Computer Vision
  (ECCV)}, pages 683--698, 2018.

\bibitem{liu2016ssd}
Wei Liu, Dragomir Anguelov, Dumitru Erhan, Christian Szegedy, Scott Reed,
  Cheng-Yang Fu, and Alexander~C Berg.
\newblock Ssd: Single shot multibox detector.
\newblock In {\em European conference on computer vision}, pages 21--37.
  Springer, 2016.

\bibitem{liu2022video}
Ze Liu, Jia Ning, Yue Cao, Yixuan Wei, Zheng Zhang, Stephen Lin, and Han Hu.
\newblock Video swin transformer.
\newblock In {\em Proceedings of the IEEE/CVF Conference on Computer Vision and
  Pattern Recognition}, pages 3202--3211, 2022.

\bibitem{ma2022robust}
Wufei Ma, Angtian Wang, Alan Yuille, and Adam Kortylewski.
\newblock Robust category-level 6d pose estimation with coarse-to-fine
  rendering of neural features.
\newblock {\em arXiv preprint arXiv:2209.05624}, 2022.

\bibitem{mao2021tfpose}
Weian Mao, Yongtao Ge, Chunhua Shen, Zhi Tian, Xinlong Wang, and Zhibin Wang.
\newblock Tfpose: Direct human pose estimation with transformers.
\newblock {\em arXiv preprint arXiv:2103.15320}, 2021.

\bibitem{marchand2015pose}
Eric Marchand, Hideaki Uchiyama, and Fabien Spindler.
\newblock Pose estimation for augmented reality: a hands-on survey.
\newblock {\em IEEE transactions on visualization and computer graphics},
  22(12):2633--2651, 2015.

\bibitem{panteleris2022pe}
Paschalis Panteleris and Antonis Argyros.
\newblock Pe-former: Pose estimation transformer.
\newblock In {\em International Conference on Pattern Recognition and
  Artificial Intelligence}, pages 3--14. Springer, 2022.

\bibitem{park2021dprost}
Jaewoo Park and Nam~Ik Cho.
\newblock Dprost: 6-dof object pose estimation using space carving and dynamic
  projective spatial transformer.
\newblock {\em arXiv preprint arXiv:2112.08775}, 2021.

\bibitem{paszke2017automatic}
Adam Paszke, Sam Gross, Soumith Chintala, Gregory Chanan, Edward Yang, Zachary
  DeVito, Zeming Lin, Alban Desmaison, Luca Antiga, and Adam Lerer.
\newblock Automatic differentiation in {PyTorch}.
\newblock In {\em NIPS Autodiff Workshop}, 2017.

\bibitem{pavlakos20176}
Georgios Pavlakos, Xiaowei Zhou, Aaron Chan, Konstantinos~G Derpanis, and
  Kostas Daniilidis.
\newblock 6-dof object pose from semantic keypoints.
\newblock In {\em 2017 IEEE international conference on robotics and automation
  (ICRA)}, pages 2011--2018. IEEE, 2017.

\bibitem{peng2019pvnet}
Sida Peng, Yuan Liu, Qixing Huang, Xiaowei Zhou, and Hujun Bao.
\newblock Pvnet: Pixel-wise voting network for 6dof pose estimation.
\newblock In {\em Proceedings of the IEEE Conference on Computer Vision and
  Pattern Recognition (CVPR)}, pages 4561--4570, 2019.

\bibitem{Rad2017BB8AS}
Mahdi Rad and Vincent Lepetit.
\newblock Bb8: A scalable, accurate, robust to partial occlusion method for
  predicting the 3d poses of challenging objects without using depth.
\newblock {\em Proc. of the IEEE International Conference on Computer Vision
  (ICCV)}, pages 3848--3856, 2017.

\bibitem{GPT2}
Alec Radford, Jeffrey Wu, Rewon Child, David Luan, Dario Amodei, and Ilya
  Sutskever.
\newblock Language models are unsupervised multitask learners.
\newblock {\em OpenAI blog}, 1(8):9, 2019.

\bibitem{rothganger20063d}
Fred Rothganger, Svetlana Lazebnik, Cordelia Schmid, and Jean Ponce.
\newblock 3d object modeling and recognition using local affine-invariant image
  descriptors and multi-view spatial constraints.
\newblock {\em International journal of computer vision}, 66(3):231--259, 2006.

\bibitem{saxena2008robotic}
Ashutosh Saxena, Justin Driemeyer, and Andrew~Y Ng.
\newblock Robotic grasping of novel objects using vision.
\newblock {\em The International Journal of Robotics Research (IJRR)},
  27(2):157--173, 2008.

\bibitem{song2020hybridpose}
Chen Song, Jiaru Song, and Qixing Huang.
\newblock Hybridpose: 6d object pose estimation under hybrid representations,
  2020.

\bibitem{sun2022onepose}
Jiaming Sun, Zihao Wang, Siyu Zhang, Xingyi He, Hongcheng Zhao, Guofeng Zhang,
  and Xiaowei Zhou.
\newblock Onepose: One-shot object pose estimation without cad models.
\newblock In {\em Proceedings of the IEEE/CVF Conference on Computer Vision and
  Pattern Recognition}, pages 6825--6834, 2022.

\bibitem{tekin2018real}
Bugra Tekin, Sudipta~N Sinha, and Pascal Fua.
\newblock Real-time seamless single shot 6d object pose prediction.
\newblock In {\em Proceedings of the IEEE Conference on Computer Vision and
  Pattern Recognition (CVPR)}, pages 292--301, 2018.

\bibitem{Tremblay2018DeepOP}
Jonathan Tremblay, Thang To, Balakumar Sundaralingam, Yu Xiang, Dieter Fox, and
  Stanley~T. Birchfield.
\newblock Deep object pose estimation for semantic robotic grasping of
  household objects.
\newblock In {\em Conference on Robot Learning (CoRL)}, 2018.

\bibitem{tulsiani2015viewpoints}
Shubham Tulsiani and Jitendra Malik.
\newblock Viewpoints and keypoints.
\newblock In {\em Proceedings of the IEEE Conference on Computer Vision and
  Pattern Recognition (CVPR)}, pages 1510--1519, 2015.

\bibitem{vaswani2017attention}
Ashish Vaswani, Noam Shazeer, Niki Parmar, Jakob Uszkoreit, Llion Jones,
  Aidan~N Gomez, {\L}ukasz Kaiser, and Illia Polosukhin.
\newblock Attention is all you need.
\newblock {\em Advances in neural information processing systems}, 30, 2017.

\bibitem{wang2019densefusion}
Chen Wang, Danfei Xu, Yuke Zhu, Roberto Mart{\'\i}n-Mart{\'\i}n, Cewu Lu, Li
  Fei-Fei, and Silvio Savarese.
\newblock Densefusion: 6d object pose estimation by iterative dense fusion.
\newblock {\em Proceedings of the IEEE Conference on Computer Vision and
  Pattern Recognition (CVPR)}, 2019.

\bibitem{wang2021gdr}
Gu Wang, Fabian Manhardt, Federico Tombari, and Xiangyang Ji.
\newblock Gdr-net: Geometry-guided direct regression network for monocular 6d
  object pose estimation.
\newblock In {\em Proceedings of the IEEE/CVF Conference on Computer Vision and
  Pattern Recognition}, pages 16611--16621, 2021.

\bibitem{wang2019normalized}
He Wang, Srinath Sridhar, Jingwei Huang, Julien Valentin, Shuran Song, and
  Leonidas~J Guibas.
\newblock Normalized object coordinate space for category-level 6d object pose
  and size estimation.
\newblock In {\em Proceedings of the IEEE Conference on Computer Vision and
  Pattern Recognition}, pages 2642--2651, 2019.

\bibitem{wen2021bundletrack}
Bowen Wen and Kostas Bekris.
\newblock Bundletrack: 6d pose tracking for novel objects without instance or
  category-level 3d models.
\newblock In {\em 2021 IEEE/RSJ International Conference on Intelligent Robots
  and Systems (IROS)}, pages 8067--8074. IEEE, 2021.

\bibitem{wen2020se}
Bowen Wen, Chaitanya Mitash, Baozhang Ren, and Kostas~E Bekris.
\newblock se (3)-tracknet: Data-driven 6d pose tracking by calibrating image
  residuals in synthetic domains.
\newblock In {\em 2020 IEEE/RSJ International Conference on Intelligent Robots
  and Systems (IROS)}, pages 10367--10373. IEEE, 2020.

\bibitem{xiang2018posecnn}
Yu Xiang, Tanner Schmidt, Venkatraman Narayanan, and Dieter Fox.
\newblock Posecnn: A convolutional neural network for 6d object pose estimation
  in cluttered scenes.
\newblock In {\em Proceedings of the European Conference on Computer Vision
  (ECCV)}, 2018.

\bibitem{yen2021inerf}
Lin Yen-Chen, Pete Florence, Jonathan~T Barron, Alberto Rodriguez, Phillip
  Isola, and Tsung-Yi Lin.
\newblock inerf: Inverting neural radiance fields for pose estimation.
\newblock In {\em 2021 IEEE/RSJ International Conference on Intelligent Robots
  and Systems (IROS)}, pages 1323--1330. IEEE, 2021.

\bibitem{yu2020ernie}
Fei Yu, Jiji Tang, Weichong Yin, Yu Sun, Hao Tian, Hua Wu, and Haifeng Wang.
\newblock Ernie-vil: Knowledge enhanced vision-language representations through
  scene graph.
\newblock {\em arXiv preprint arXiv:2006.16934}, 2020.

\bibitem{zakharov2019dpod}
Sergey Zakharov, Ivan Shugurov, and Slobodan Ilic.
\newblock Dpod: 6d pose object detector and refiner.
\newblock In {\em Proceedings of the IEEE International Conference on Computer
  Vision}, pages 1941--1950, 2019.

\bibitem{zhao2022dpit}
Shuaitao Zhao, Kun Liu, Yuhang Huang, Qian Bao, Dan Zeng, and Wu Liu.
\newblock Dpit: Dual-pipeline integrated transformer for human pose estimation.
\newblock {\em arXiv preprint arXiv:2209.02431}, 2022.

\bibitem{zheng20213d}
Ce Zheng, Sijie Zhu, Matias Mendieta, Taojiannan Yang, Chen Chen, and Zhengming
  Ding.
\newblock 3d human pose estimation with spatial and temporal transformers.
\newblock In {\em Proceedings of the IEEE/CVF International Conference on
  Computer Vision}, pages 11656--11665, 2021.

\bibitem{zhu2020deformable}
Xizhou Zhu, Weijie Su, Lewei Lu, Bin Li, Xiaogang Wang, and Jifeng Dai.
\newblock Deformable detr: Deformable transformers for end-to-end object
  detection.
\newblock {\em arXiv preprint arXiv:2010.04159}, 2020.

\end{thebibliography}
}
\clearpage
%%%%%%%%%%%%%%%%%%%%%%%%%%%%%%%%%%%%%%%%%%%%%%%%%%%%%%%%%%%%
% \input{checklist}
\clearpage
\appendix
\section{Appendix}
\label{sec:appendix}
\subsection{Code Release:}
Training code and pretrained models are available at \href{https://github.com/ApoorvaBeedu/VideoPose}{https://github.com/ApoorvaBeedu/VideoPose}.

\subsection{Baseline implementation:}
\begin{figure*}[]
    \centering
    \includegraphics[width=0.9\textwidth]{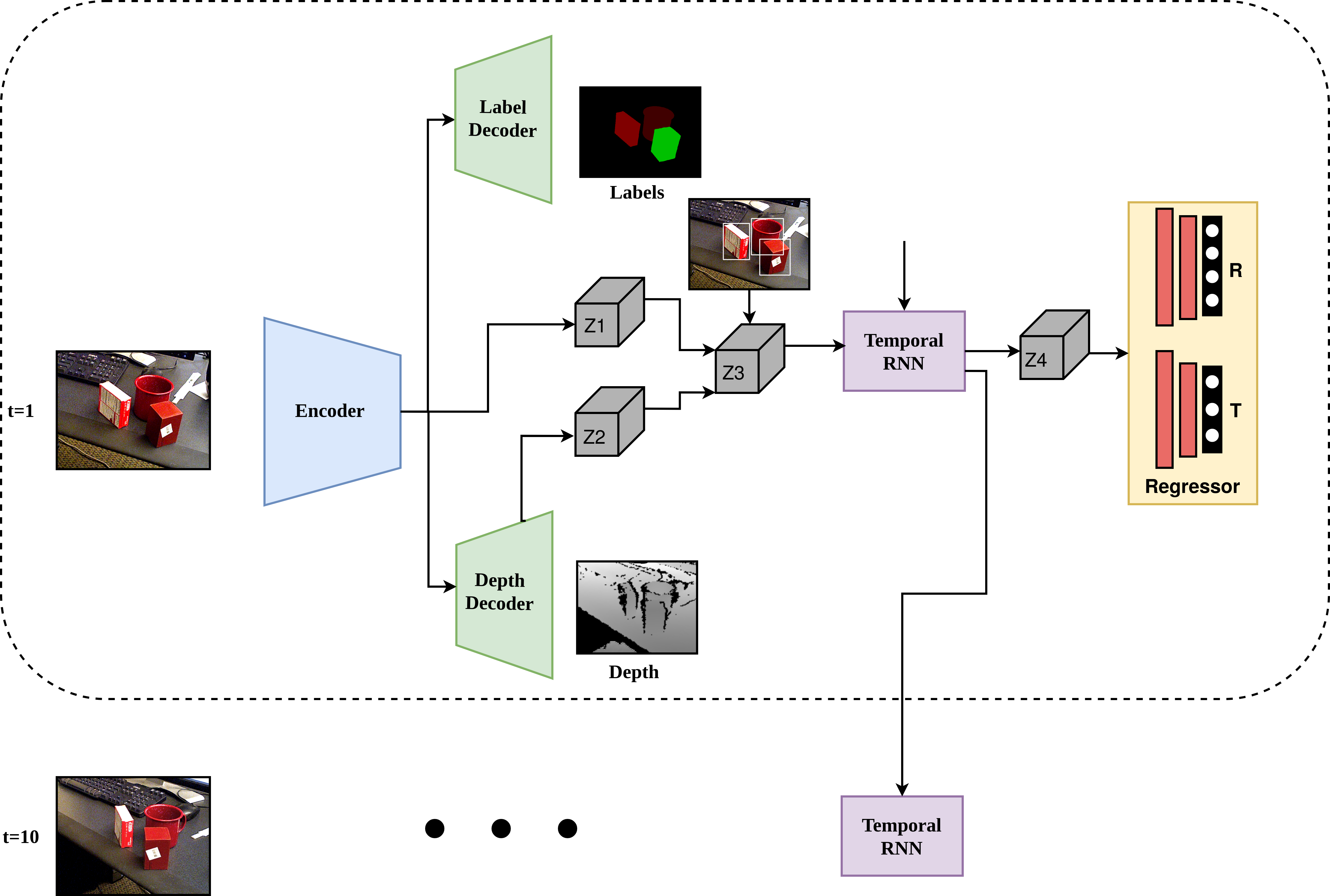} \vspace{.1in} 
    \caption{Overview of our VideoPose framework for 6D object pose estimation baseline. We use the same encoder as in ~\cite{xiang2018posecnn}. }
    \label{fig:overview_convgru}
\end{figure*}

As mentioned in Section~\ref{sec:experiments}, to establish a baseline for video related pose estimation, we implemented a ConvGRU based temporal encoder as our baseline. 
In order to improve performance and match the pipeline of PoseCNN, several modifications were made to the baseline framework that does not exist in our Transformer based model. 
The overview diagram for ConvGRU based implementation can be seen in Figure~\ref{fig:overview_convgru}. For the baseline, we train a depth decoder, and concatenate the features with the image features before feeding it into the ConvGRU. The $\boldsymbol{[R | T]}$ regressor is the same as in the main framework.

\subsubsection{Baseline Training Statergy:}
In addition to the losses defined in Section~\ref{sec:training_stratergy}, two additional losses were used in the training of ConvGRU baseline. 
We use the $L_1$ loss to learn depth ($L_\text{depth}$), and cross entropy loss for semantic segmentation ($L_\text{label}$). 
The total loss can be defined as:
\begin{equation} \label{eq:loss2}
    L(\mathbf{\widetilde{q}}, \mathbf{q}, \mathbf{\widetilde{t}}, \mathbf{t}) = L_\text{depth} + L_\text{label} + L_\text{pose} + L_\text{reg} + L_\text{inner\_prod}.
\end{equation} 

\end{document}